\def\year{2020}\relax
\documentclass[letterpaper]{article} 
\usepackage{aaai20}  
\usepackage{times}  
\usepackage{helvet} 
\usepackage{courier}  
\usepackage[hyphens]{url}  
\usepackage{graphicx} 
\usepackage{graphicx}  
\usepackage{amsmath}
\usepackage{amsfonts}
\usepackage{amssymb}

\usepackage{multirow}

\title{JSNet: Joint Instance and Semantic Segmentation of 3D Point Clouds}
\author{Lin Zhao, Wenbing Tao\textsuperscript{}\thanks{Corresponding author.}\\
National Key Laboratory of Science and Technology on Multispectral Information Processing \\
School of Artifical Intelligence and Automation, Huazhong University of Science and Technology, China\\
\{linzhao, wenbingtao\}@hust.edu.cn 
}
\begin{document}

\maketitle

\begin{abstract}
In this paper, we propose a novel joint instance and semantic segmentation  approach, which is called JSNet, in order to address the instance and semantic segmentation of 3D point clouds simultaneously. Firstly, we build an  effective backbone network to extract robust features from the raw point clouds. Secondly, to obtain more discriminative features, a point cloud feature fusion module is proposed to fuse the different layer features of the backbone network. Furthermore, a joint instance semantic segmentation module is developed to transform semantic features into instance embedding space, and then the transformed features are further fused with instance features to facilitate instance segmentation. Meanwhile, this module also aggregates instance features into semantic feature space to promote semantic segmentation. Finally, the instance predictions are generated by applying a simple mean-shift clustering on instance embeddings. As a result, we evaluate the proposed JSNet on a large-scale 3D indoor point cloud dataset S3DIS and a part dataset ShapeNet, and compare it with existing approaches. Experimental results demonstrate our approach outperforms the state-of-the-art method in 3D instance segmentation with a significant improvement in 3D semantic prediction and our method is also beneficial for part segmentation. The source code for this work is available at \url{https://github.com/dlinzhao/JSNet}. 
\end{abstract}

\section{Introduction}
Semantic segmentation is the task which is used to segment all informative regions in a scene and classify each region into a specific class. Instance segmentation is different from semantic segmentation for that different objects of the same class will have different labels. Both the two tasks have a wide applications in real-world scenarios, e.g., autonomous driving and mobile-based navigation. In 2D images, those two tasks have achieved remarkable results \cite{deeplabv3plus2018,he2017mask,li2018attention}. However, the studies of 3D semantic and instance segmentation are still facing huge challenges, e.g., large-scale with noisy data processing, high computation as well as memory consumption.

Literature research shows that 3D scene data have different representations, e.g., volumetric grids \cite{Wu_2015_CVPR,Nguyen_2016_CVPR,maturana2015voxnet} and 3D point clouds \cite{qi2017pointnet++,li2018pointcnn,wang2018dynamic,yu2018pu}. Compared with other representations, point cloud is a more compact and intuitive representation of 3D scene data. Recently, more efficient and powerful deep learning network architectures \cite{qi2017pointnet++,wu2018pointconv,Li_2018_CVPR} have been proposed to directly process point clouds and shown promising results in point cloud classification and part segmentation. Those approaches are often used as feature extraction network in other tasks, e.g., instance segmentation and semantic segmentation.

\begin{figure}[t]
	\centering
	\includegraphics[width=0.9\columnwidth]{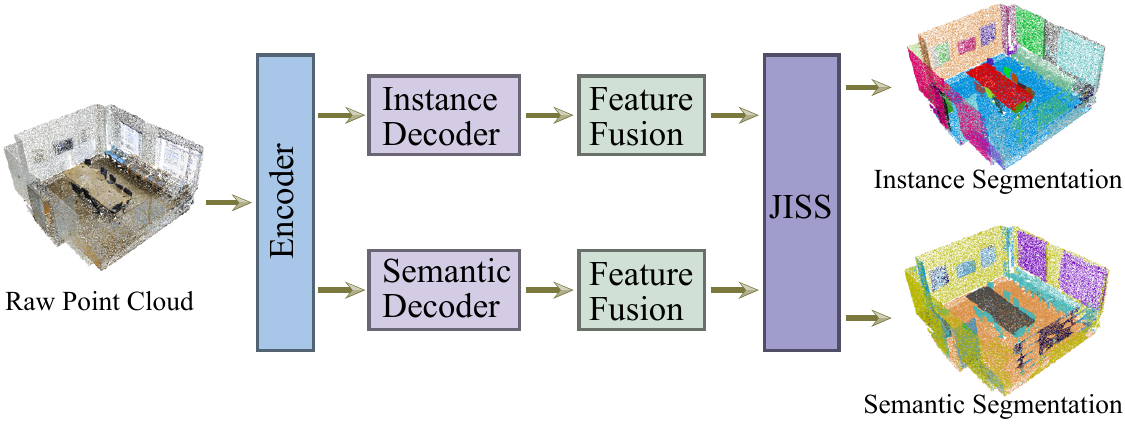}
	\caption{Our network JSNet takes raw point clouds as inputs and gets outputs instance and semantic segmentation results for each point. JISS stands for Joint Instance and Semantic Segmentation.}
	\label{fig:network_framework}
\end{figure}

In previous works, instance segmentation and semantic segmentation have often been processed respectively or instance segmentation is treated as a post-processing task of semantic segmentation \cite{Wang_2018_CVPR,pham2019real}. However, those two problems are related because points of different categories belong to different instances and points of the same instance belong to the same class. Recently, \cite{Pham_2019_CVPR} handles the two problems with multi-task pointwise network and multi-value Conditional Random Field (CRF). However, the CRF is an individual part behind the Convolutional Neural Network (CNN), it is difficult to explore the performance of their combination. Moreover, this method does not investigate whether semantic segmentation and instance segmentation can facilitate each other. At the same time, ASIS \cite{Wang_2019_CVPR} is proposed to address the two tasks simultaneously, which adapts  the semantic feature to instance feature space by a fully connected layer and aggregates instance feature to semantic feature space by $K$ Nearest Neighbor (kNN). However, the performance of this method is limited because it is difficult to choose the right $K$ value and distance metric for kNN. Besides, it has expensive computation and memory consumption because it will generate a high order sparse tensor during training process. 

In this work, we introduce a joint instance semantic segmentation neural network of 3D point clouds called JSNet to address the two fundamental problems: semantic segmentation and instance segmentation. The proposed network JSNet includes four parts: a shared feature encoder, two parallel branch decoders, a feature fusion module for each decoder, a joint segmentation module. The feature encoder and decoders are built based on PointNet++ \cite{qi2017pointnet++} and PointConv \cite{wu2018pointconv} to learn more effective high-level semantic features. To obtain more discriminative features, we propose a point cloud feature fusion module to fuse the high-level and low-level information to refine the output features. In order to make the two tasks promote each other, a novel joint instance and semantic segmentation module is proposed to handle instance and semantic segmentation simultaneously. Specifically,this module transforms semantic features into instance embedding space by a 1D convolution and then the transformed features are further fused with instance features to facilitate instance segmentation. Meanwhile, this module also aggregates instance features into semantic feature space by implicit learning to promote semantic segmentation. Thus, our approach can be used to learn instance-aware semantic fusion features and semantic-aware instance embedding features, which can make the predictions of those points more accurate.

To summarize, the main contributions of our work are as follows:

\begin{itemize}
\item We design a more efficient Point Cloud Feature Fusion (PCFF) module to generate more discriminative features and improve the accuracy of point predictions.
\item We propose a novel Joint Instance and Semantic Segmentation (JISS) module to make instance segmentation and semantic segmentation mutual promote. This module further improve the accuracy with acceptable GPU memory consumption during training process.
\item We achieve impressive results on the S3DIS dataset \cite{Armeni_2016_CVPR} along with a significant improvement on the 3D instance segmentation task. Additionally, our experiments on the ShapeNet dataset \cite{yi2016scalable} indicate that JSNet also can achieve satisfactory performance for part segmentation task.
\end{itemize}

\section{Related Work}
In this section, we briefly review some point cloud feature extraction works, and some existing approaches for semantic and instance segmentation in 3D scene. Especially, we concentrate on deep neural network-based methods applied to 3D point clouds because of their proven robustness and efficiency in the field.

\subsection{Deep learning for 3D Point Clouds}
Although deep learning has been successfully used for 2D images, there are still many challenges in the feature learning capabilities of 3D point clouds with irregular data structures. Recently, PointNet \cite{qi2016pointnet} is one of the first approaches of directly applying neural networks to point clouds. It uses shared Multi-Layer Perceptron (MLP) and max pooling to learn deep features from unordered point sets. However, PointNet has difficulty in capturing local region features. This drawback has been addressed by PointNet++ \cite{qi2017pointnet++} with a hierarchical neural network. The max pooling operation is a key structure to extract features from points for both PointNet and PointNet++. But it only keeps the strongest activation on a local or global region of feature maps, which may cause some useful detailed information lost for semantic and instance segmentation tasks.

Some later works \cite{Simonovsky_2017_CVPR,hermosilla2018monte,xu2018spidercnn} extract features of point clouds with learning continuous filters for convolution calculation. The work \cite{Simonovsky_2017_CVPR} firstly presents the idea that learning continuous filters with edge-conditioned into 3D graph. Furthermore, Dynamic graph CNN \cite{wang2018dynamic} introduces a method to update the graph dynamically. The following work PointConv \cite{wu2018pointconv} proposes an inverse density scale to re-weight the continuous function learned by MLP and compensate the non-uniform sampling, while it also needs high GPU memory during training process.

\subsection{Semantic\&Instance Segmentation on Point Clouds}
For semantic segmentation, methods \cite{Zhao_2017_CVPR,deeplabv3plus2018} based on full convolutional networks \cite{Long_2015_CVPR} have achieved tremendous progress in 2D domain. As for 3D semantic segmentation, 3D-FCNN introduced by \cite{huang2016point} predicts a coarse voxel label with a 3D fully convolutional neural network. SEGCloud \cite{tchapmi2017segcloud} extends 3D-FCNN with trilinear interpolation and fully connected conditional random fields. RSNet \cite{huang2018recurrent} models local dependencies for point clouds with a slice pooling layer, Recurrent Neural Network (RNN) layers, and a slice unpooling layer. 3P-RNN \cite{ye20183d} models the inherent contextual features for semantic segmentation by using a pointwise pyramid pooling module and explores long-range spatial dependencies with two-direction hierarchical RNNs. Recently, GAC \cite{wang2019graph}, a graph attention convolution, is proposed to capture the structured feature of point clouds with dynamical kernels to adapt the structure of an object. However, there are few previous works which focus on semantics segmentation by using the advantages of instance embedding.

For instance segmentation, approaches \cite{li2018attention,huang2019msrcnn} based on Mask R-CNN \cite{he2017mask} dominate it on 2D images. However, there are few studies for 3D instance segmentation. SGPN \cite{Wang_2018_CVPR} generates instance proposals from learning a similarity matrix of the point features with a double-hinge loss. GSPN \cite{yi2019gspn} generates proposals by reconstructing shapes and outputs the final segmentation results based on PointNet++. 3D-BoNet \cite{yang2019learning} directly regresses 3D bounding boxes and predicts point-level masks for all instances simultaneously. Similarly, there are few works which segment instances using the advantages of semantic fusion.

However, most of the previous works tackle the two tasks separately. Very recently, \cite{Pham_2019_CVPR} proposes a multi-task pointwise network (MT-PNet) for predicting the semantic categories and instance embedding vectors and then uses a multi-value conditional random field (MV-CRF) as a post-processing. However, the CRF is an individual part behind the CNN, and it is difficult to explore the performance of their combination. Moreover, this method does not investigate whether semantic segmentation and instance segmentation can promote each other. Therefore, the performance improvement is not obvious. Meanwhile, ASIS \cite{Wang_2019_CVPR} is proposed to segment instances and semantics for 3D point clouds at once, which uses PointNet or PointNet++ as backbone network and then concatenates a proposed module ASIS. The ASIS adapts the semantic features to instance feature space by a fully connected layer and aggregates instance features to semantic feature space by kNN. While the approach \cite{Wang_2019_CVPR} has difficult to choose the right $K$ value and distance metric for kNN, and it also has high memory cost because it will produce a high order sparse matrix at training process.

\section{Proposed Method}
\begin{figure*}[t]
	\centering
	\includegraphics[width=0.88\textwidth]{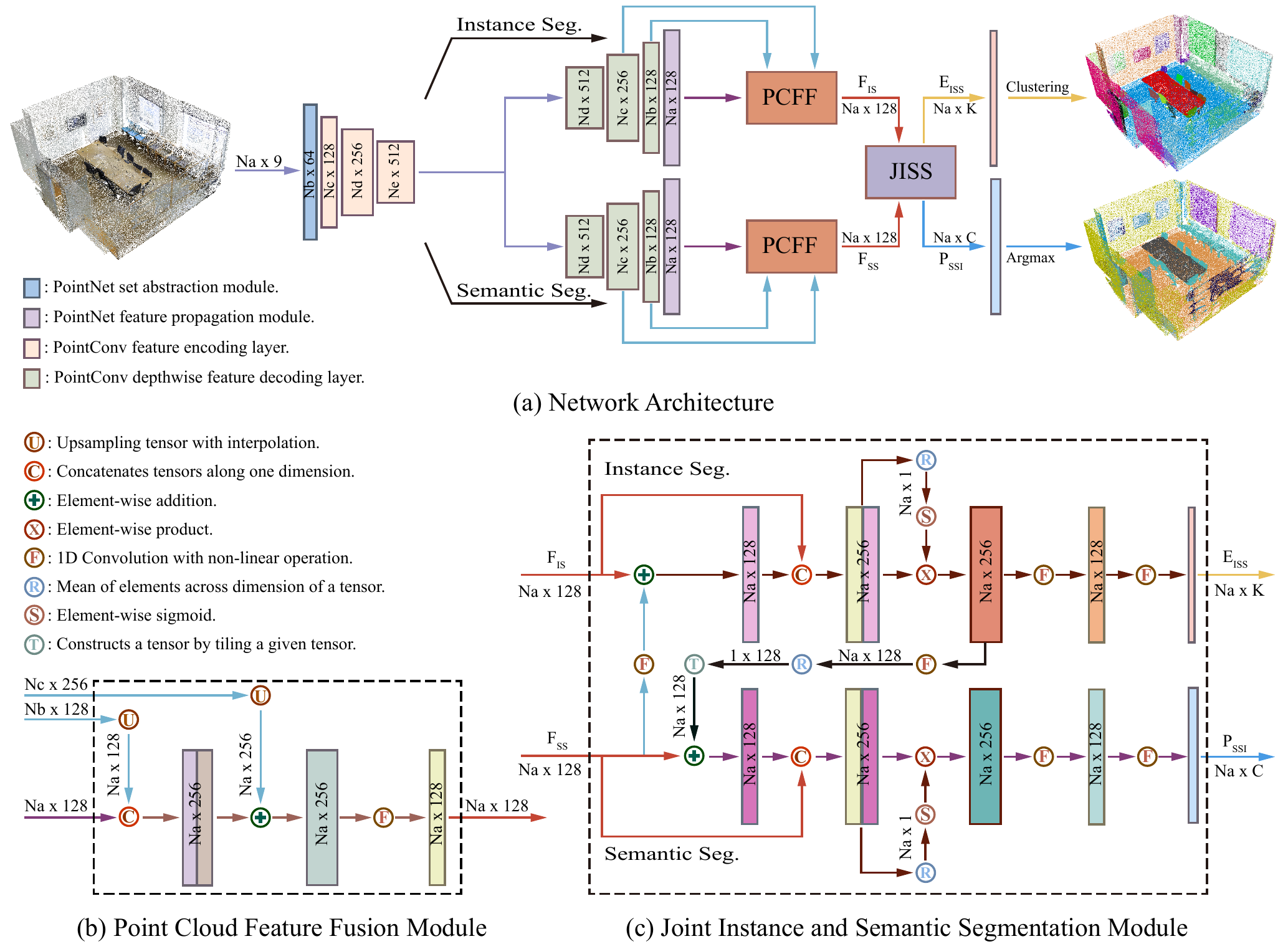}
	\caption{An overview of the Joint Instance Semantic Segmentation Neural Network of 3D Point Cloud (JSNet). (a) Illustration of the network architecture. (b) Components of the Point Cloud Feature Fusion (PCFF) module. (c) Components of the Joint Instance and Semantic Segmentation (JISS) module. Different colored blocks represent different modules in (a), while those blocks represent different features in (b) and (c).}
	\label{fig:network_framework_detail}
\end{figure*}

In this section, firstly, we describe the whole network architecture of our proposed JSNet for instance and semantic segmentation of 3D point clouds. Then, we elaborate on the two main components of our proposed network, including the Point Cloud Feature Fusion (PCFF) module and the Joint Instance and Semantic Segmentation (JISS) module, respectively. 

\subsection{Network Architecture}
The whole network illustrated in Figure \ref{fig:network_framework_detail}(a) composes with four main components including a shared encoder, two parallel decoders, a point cloud feature fusion module for each decoder, a joint segmentation module as the last part. For the two parallel branches, one aims to extract semantic feature for each point, while the other one is for instance segmentation task. Specifically for the feature encoder and two decoders, we can directly use PointNet++ or PointConv as our backbone network by duplicating a decoder because the two decoders have the same structure. However, as is mentioned above, as for instance or semantic segmentation, the PointNet++ may lose detailed information because of max pooling operation and the PointConv has expensive GPU memory consumption during training process. In this work, we combine the PointNet++ and PointConv to build a more effective backbone network with acceptable memory cost. The encoder of the backbone is built by concatenating a set abstraction module of PointNet++ and three feature encoding layers of PointConv. Similarly, the decoders are composed with three depthwise feature decoding layers of PointConv followed by a feature propagation module of PointNet++.

For the whole pipeline, our network takes a point cloud of size $N_{a}$ as input, then encodes it into a $N_e \times 512$ shaped matrix by the shared feature encoder. Next, the output of feature encoder is input into the two parallel decoders and processed by their following components separately. The semantic branch decodes the shared features and fuses the features of different layers into a semantic feature matrix $F_{SS}$ shaped with $N_a \times 128$. Similarly,  the instance branch outputs an instance feature matrix $F_{IS}$ after the PCFF module. Finally, both the semantic features and the instance features are fetched and processed  by the JISS module, and then output two feature matrices. One of the matrices $P_{SSI}$ shaped with $N_{a} \times C$ which is used to predict the semantic categories, where $C$ is the number of semantic categories. The other one $E_{ISS}$ shaped with $N_a \times K$ is an instance feature matrix and it is used to predict the instance labels for each point, where $K$ is the dimension of the embedding vector. In the embedding space, the embeddings represent the instance relationship of points: the points belonging to the same instance object are close, and the points of the different instances are kept away from each other.

At training time, the loss function $\mathcal{L}$ of our network consists of semantic segmentation loss $\mathcal{L}_{sem}$ and instance embedding loss $\mathcal{L}_{ins}$:
\begin{equation}
	\mathcal{L} = \mathcal{L}_{sem} + \mathcal{L}_{ins},
\end{equation}
where $\mathcal{L}_{sem}$ is defined with the classical cross entropy loss. As for the instance embedding loss, we utilize a discriminative function to express the embedding loss $\mathcal{L}_{ins}$ inspired by the work in \cite{de2017semantic}. Specifically, the instance embedding loss function is formulated as follows:
\begin{equation}
	\mathcal{L}_{ins} = \mathcal{L}_{pull} + \mathcal{L}_{push}, 
\end{equation}
where $\mathcal{L}_{pull}$ pulls embeddings close to the mean embedding of instance, while the $\mathcal{L}_{push}$ makes the mean embedding of different instances seperated from each other. Given the number of instances $M$, the number of elements $N_m$ in the $m$-th instance, the embedding $e_n$ of point, and the mean of embeddings $\mu_{m}$ in the $m$-th instance. Each term is rewritten as follows:
\begin{equation}
	\mathcal{L}_{pull} = \frac{1}{M}\sum_{m=1}^{M}\frac{1}{N_{m}}\sum_{n=1}^{N_{m}}\left[\left\| \mu_{m} - e_{n} \right\|_{1} - \delta_{v} \right]_{+}^{2},
\end{equation}
\begin{equation}
	\mathcal{L}_{push} = \frac{1}{M\left(M - 1\right)}\underset{ i \ne j }{ \sum_{i=1}^{M} \sum_{j=1}^{M} } \left[2\delta_{d} - \left\| \mu_{i} - \mu_{j} \right\|_{1} \right]_{+}^{2},
\end{equation}
where $\left[x\right]_{+} = \max\left(0, x\right)$; $\left\|\cdot\right\|_{1}$ is $L_{1}$ distance; $\delta_{v}$ and $\delta_{d}$ are margins for $\mathcal{L}_{pull}$ and $\mathcal{L}_{push}$ respectively.

At testing time, the final instance labels are generated by using a simple mean-shift clustering \cite{comaniciu2002mean} on the embeddings and the final semantic categories are obtained by using a argmax operation.

\subsection{Point Cloud Feature Fusion Module}
In the segmentation and detection tasks for 2D image, only the feature of last layer is used for prediction in previous works, while different layer features are fused in subsequent approaches  \cite{Lin_2017_CVPR,he2017mask,deeplabv3plus2018} because the high level layer has richer semantic information while the low level has much more detailed information. Those works indicate that the fused features are beneficial for better prediction.

Based on observation above, we propose a Point Cloud Feature Fusion (PCFF) module for semantic and instance segmentation in point clouds. Figure \ref{fig:network_framework_detail}(b) presents the details of the structure. Considering the precision, computation and GPU memory consumption, we only fuse the last three layers of the decoder. We use $F_{a}$, $F_{b}$ and $F_{c}$ to represent those feature matrices of the decoder with shape $N_{a} \times 128$, $N_{b} \times 128$ and $N_{c} \times 256$ respectively. Firstly, we concatenate $F_{a}$ and $F_{b}^{'}$ upsampling with interpolation from $F_{b}$. Then the former output is added to $F_{c}^{'}$ (upsampling from $F_{c}$) element-wise and a convolution is applied to the previous result. Following \cite{qi2017pointnet++}, the interpolation is achieved by using an inverse square distance weighted average based on three nearest neighbors. Finally, the PCFF generates a fused feature matrix with shaped $N_{a} \times 128$. This module can refine the output features from the decoder with acceptable computation and memory consumption.

\subsection{Joint Instance and Semantic Segmentation}
In fact, both the semantic segmentation and the instance segmentation map the initial point cloud features to different new high-level feature spaces separately. In the semantic feature space, points of the same semantic category are clustered together, while the different classes are separated. In the instance feature space, points of the same instance object are closely assembled, while points of different instances are separated. It indicates that we could extract semantic awareness information from the semantic feature space to integrate the information  into the instance features and generates semantic-aware instance embedding features, and vice versa.

Based on this observation, we propose a Joint Instance Semantic Segmentation (JISS) module to obtain semantic labels and segment instance objects simultaneously, as is illustrated in Figure \ref{fig:network_framework_detail}(c). The JISS module transforms semantic features into instance embedding space and then the transformed features are further fused with instance features to facilitate instance segmentation. Meanwhile, this module also aggregates instance features into semantic feature space to promote semantic segmentation. Specifically, the semantic feature matrix $F_{SS}$ is transformed into instance feature space as $F_{SST}$ by a 1D convolution (Conv1D), and the $F_{SST}$ is added to instance feature matrix $F_{IS}$ element-wise as $F_{ISS}$. Then, we model the spatial correlation of point features to enhance important features by concatenating the feature $F_{IS}$ and $F_{ISS}$ into a $F_{ISSC}$, and then the $F_{ISSC}$ is applied a mean of elements across dimension (Mean) and an element-wise sigmoid (Sigmoid) to generate a weight matrix $F_{ISR}$. Finally, the feature matrix $F_{ISSC}$ multiply the $F_{ISR}$ to generate the feature matrix $F_{ISSR}$  followed by two 1D convolution to produce the instance embedding feature $E_{ISS}$ shaped with $N_{a} \times K$. The process can be formulated as follows:

\begin{equation}
	F_{ISSC} = Concat\left( F_{IS}, F_{IS} + Conv1D\left(F_{SS}\right) \right),
\end{equation}
\begin{equation}
	F_{ISSR} = F_{ISSC} \cdot Sigmoid ( Mean( F_{ISSC} ) ),
\end{equation}
\begin{equation}
	E_{ISS} = Conv1D\left( Conv1D\left( F_{ISSR} \right) \right),
\end{equation}
where instance embedding feature matrix $E_{ISS}$ is used to generate final instance labels by using mean-shift clustering.

For the semantic segmentation branch, given the instance embeddings $F_{ISSR}$, this module integrates the $F_{ISSR}$ into semantic feature space as $F_{ISST}$ with a 1D convolution followed by a mean of elements across dimension and a tiling operation. Next, other operations are similar to the instance branch except the last layer which outputs an instance-aware semantic feature matrix $P_{SSI}$ shaped with $N_{a} \times C$. We also formulate this procedure as follows:
\begin{equation}
F_{ISST} = Tile\left( Mean\left(  Conv1D\left(F_{ISSR}\right) \right) \right),
\end{equation}
\begin{equation}
F_{SSI} = Concat ( F_{SS}, F_{SS} + F_{ISST} ),
\end{equation}
\begin{equation}
F_{SSIR} = F_{SSI} \cdot Sigmoid ( Mean ( F_{SSI}  )  ),
\end{equation}
\begin{equation}
P_{SSI} = Conv1D\left( Conv1D\left( F_{SSIR} \right) \right),
\end{equation}
where $F_{SSI}$ is a instance-fused feature matrix and the $F_{SSIR}$ is a feature fusion matrix for semantic segmentation. The final instance-aware semantic features are fed into the last classifier to predict the categories for each point.

\section{Experiments}
\subsection{Datasets and Evaluation Metrics}
We evaluate our approach on the following two public datasets: Stanford Large-Scale 3D Indoor Spaces (S3DIS) \cite{Armeni_2016_CVPR} and ShapeNet \cite{yi2016scalable}. The S3DIS is an indoor 3D point cloud dataset that contains six areas of three different buildings and have 272 rooms and involve 13 categories in total. For a principled evaluation, we follow the same k-fold cross validation as in \cite{qi2016pointnet}, and we also present the results of the 5-th fold (Area 5) following \cite{tchapmi2017segcloud} because Area 5 is not in the same building as other areas and there exist some differences between the objects in Area 5 and other areas. Moreover, we also evaluate our algorithm on ShapeNet dataset. This dataset contains 16881 CAD models from 16 categories annotated with 50 types of parts and the models in each category are labeled with two to five parts. We follow the official split of 795 scenes as training set, 654 scenes as testing set. The instance annotations generated following \cite{Wang_2018_CVPR} are regarded as instance ground truth labels.

For semantic segmentation evaluation, overall accuracy (oAcc), mean accuracy (mAcc) and mean IoU (mIoU) are calculated across over all the categories. For instance segmentation, we evaluate our method including mean precision (mPrec), mean recall (mRec) with IoU threshold 0.5 and (weighted) coverage (Cov, WCov) \cite{ren2017end,Wang_2019_CVPR}. The Cov scores measure the instance-wise IoU for each prediction matched with ground truth instance averaged over the scene. And then the Cov is further weighted with the size of ground-truth instances to obtain WCov. Given the predicted regions $\mathcal{P}$ and ground truth regions $\mathcal{G}$, the Cov and WCov are formulated as:
\begin{equation}
	Cov\left( \mathcal{G}, \mathcal{P} \right) = \sum_{m=1}^{\left| \mathcal{G} \right|} \frac{1}{\left| \mathcal{G} \right|} \max_{n} \text{IoU} \left( r_{m}^{\mathcal{G}}, r_{n}^{\mathcal{P}} \right),
\end{equation}
\begin{equation}
	WCov\left( \mathcal{G}, \mathcal{P} \right) = \sum_{m=1}^{\left| \mathcal{G} \right|} w_{m} \max_{n} \text{IoU} \left( r_{m}^{\mathcal{G}}, r_{n}^{\mathcal{P}} \right),
\end{equation}
\begin{equation}
	w_{m} = \frac{\left| r_{m}^{\mathcal{G}} \right|}{\sum_{k} \left| r_{k}^{\mathcal{G}} \right|},
\end{equation}
where $\left| r_{m}^{\mathcal{G}} \right|$ is the number of points in ground truth region $m$.

\subsection{Implementation Details}\label{sec:experiments-implement}
For the large scale dataset S3DIS, each point in our model is represented by a 9-dim vector (XYZ, RGB and normalized location as to the room). Following experimental settings in PointNet \cite{qi2016pointnet}, we split the rooms into overlapped blocks of area $1m \times 1m$, and each block contains 4096 points. During training process, we configure the network with $\delta_{v} = 0.5$, $\delta_{d} = 1.5$ 
and $K = 5$, where $K$ is the dimension of the embedding. We train the network for 100 epochs with batch size 24 on a single NVIDIA GTX1080Ti. We use Adam optimizer to optimize the network with momentum set to 0.9, base learning rate set to 0.001, and decay by 0.5 every $12.5k$ iterations. At test time, We use mean-shift clustering with bandwidth 0.6 to generate instance objects and merge instances of different blocks by using BlockMerging algorithm \cite{Wang_2018_CVPR}. For ShapeNet dataset, each model is sampled into a point cloud with 2048 points represented by a 6-dim vector (XYZ and normal) as in \cite{qi2017pointnet++}.

\subsection{Instance Segmentation on the S3DIS dataset}
\begin{figure}[t]
	\centering
	\includegraphics[width=0.9\columnwidth]{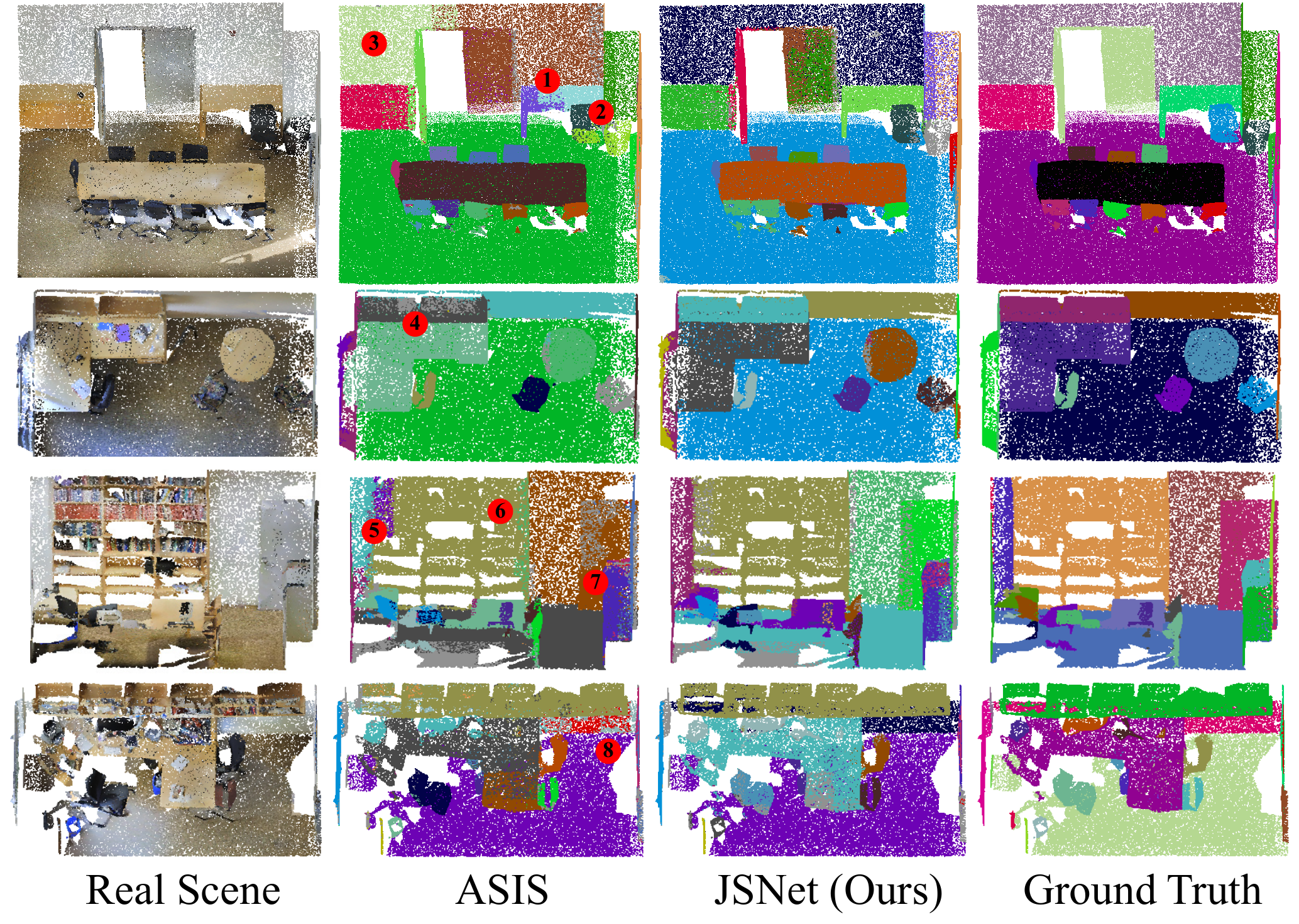} 
	\caption{Comparison results of ASIS and our method in instance segmentation task on S3DIS. Different colors represent different instances. Different numbers indicate that the segmentation results of our method are better than the ASIS in nearby area. }
	\label{fig:reulst_instance}
\end{figure}

As is depicted in Table \ref{tab:s3dis_ins}, we present the performance of our approach in instance segmentation task on S3DIS dataset. In this task, we evaluate and compare our method with exist state-of-the-art methods including SGPN \cite{Wang_2018_CVPR}, MT-PNet \cite{Pham_2019_CVPR}, MV-CRF \cite{Pham_2019_CVPR}, ASIS \cite{Wang_2019_CVPR}, 3D-BoNet \cite{yang2019learning}. We can see our network outperforms the other methods on S3DIS. Among them, ASIS is the most similar approach to our method. Compared with ASIS, our network JSNet achieves significant improvements on the four evaluation metrics. Especially on Area 5 of S3DIS, the improvements are more significant for each metric: 4.1 mCov, 3.7 mWCov, 4.5 mRec and 6.8 mPrec. Compared with the latest method 3D-BoNet on six fold experiments, our approach is also slightly better. Qualitative results are presented in Figure \ref{fig:reulst_instance}.

\begin{table}[t]
	\caption{Instance segmentation results on S3DIS dataset.}
	\smallskip
	\centering
	\resizebox{.95\columnwidth}{!}
	{
		\smallskip
		\begin{tabular}{c|c|c|c|c|c}
			\hline
			~ & Method         & mCov          & mWCov         & mRec          & mPrec \\
			\hline
			\multirow{3}{*}{5-th fold}
			  & SGPN           & 32.7          & 35.5          & 28.7          & 36.0 \\
			~ & ASIS           & 44.6          & 47.8          & 42.4          & 55.3 \\
			~ & JSNet (Ours)   & \textbf{48.7} & \textbf{51.5} & \textbf{46.9} & \textbf{62.1} \\
			\hline\hline
			\multirow{6}{*}{6 fold}
			  & SGPN           & 37.9          & 40.8          & 31.2          & 38.2 \\
			~ & MT-PNet        & -             & -             & -             & 24.9 \\
			~ & MV-CRF         & -             & -             & -             & 36.3 \\
			~ & ASIS 	       & 51.2          & 55.1          & 47.5          & 63.6 \\
			~ & 3D-BoNet       & -             & -             & 47.6          & 65.6 \\
			~ & JSNet (Ours)   & \textbf{54.1} & \textbf{58.0} & \textbf{53.9} & \textbf{66.9} \\
			\hline
		\end{tabular}
	}
	\label{tab:s3dis_ins}
\end{table}

\subsection{Semantic Segmentation on the S3DIS dataset}
\begin{figure}[t!]
	\centering
	\includegraphics[width=0.9\columnwidth]{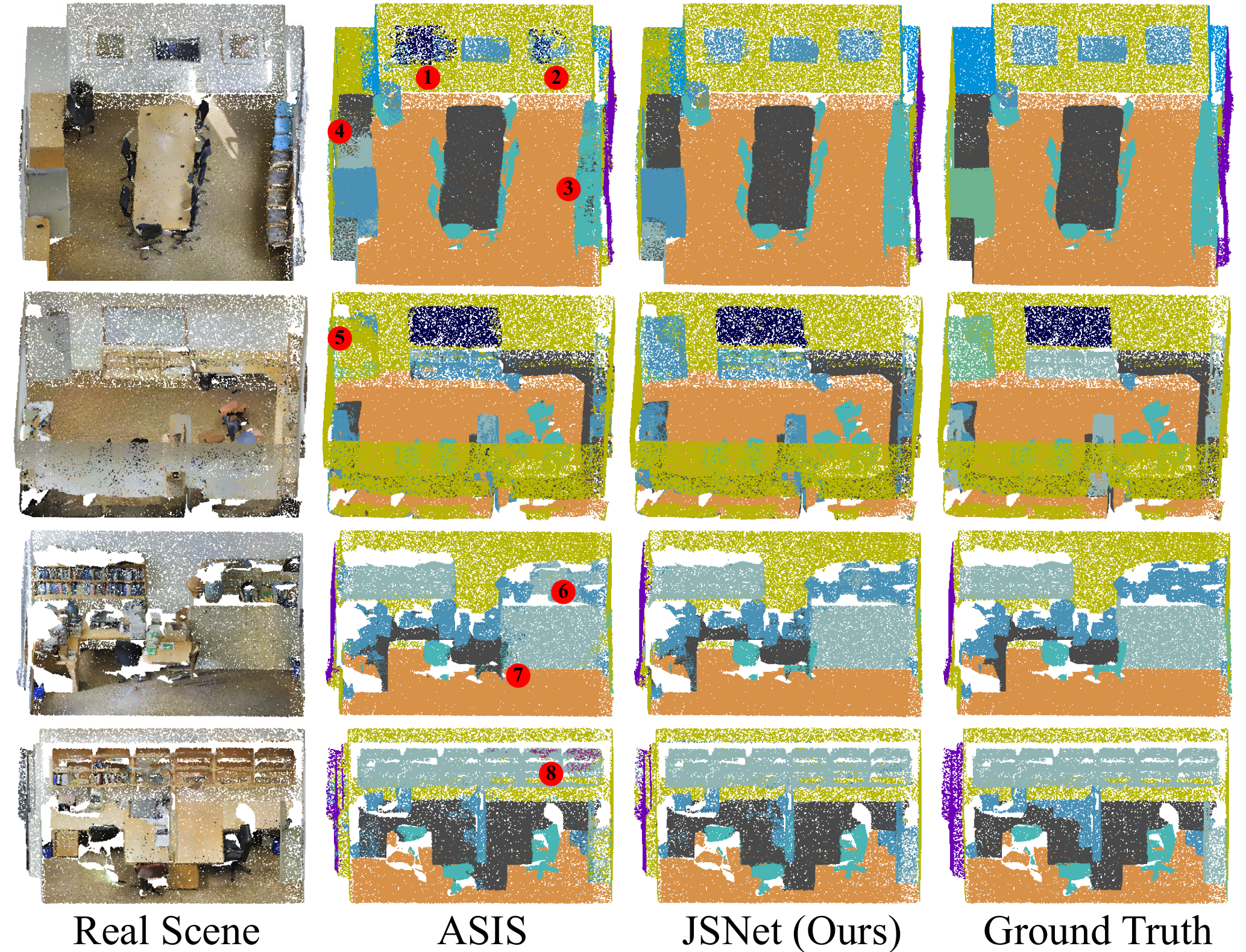}
	\caption{Comparison results of ASIS and our method in semantic segmentation task on S3DIS. Different numbers indicate that the segmentation results of our method are better than the ASIS in nearby area.}
	\label{fig:reulst_semantic}
\end{figure}

Table \ref{tab:s3dis_sem} presents the quantitative results of our architecture in semantic segmentation task on S3DIS dataset. As is seen from Table \ref{tab:s3dis_sem}, our approach outperforms the baseline PointNet \cite{qi2016pointnet} by 11.4 mAcc, 8.4 oAcc and 12.8 mIoU in over all accuracy on six fold cross validation experiments. For the generalizability evaluation on Area 5 of S3DIS,  the performance is improved with 9.3 mAcc, 4.2 oAcc and 11.1 mIoU respectively. In addition, we also compare our method with other state-of-the-art methods on 6 fold or 5-th fold of S3DIS. Our model is slightly better than SEGCloud \cite{tchapmi2017segcloud}, RSNet \cite{huang2018recurrent}, 3P-RNN \cite{ye20183d}, MT-PNet \cite{Pham_2019_CVPR}, MV-CRF \cite{Pham_2019_CVPR}, and ASIS \cite{Wang_2019_CVPR}. Qualitative results are presented in Figure \ref{fig:reulst_semantic}. 

\begin{table}[t]
	\caption{Semantic segmentation results on S3DIS dataset.}
	\smallskip
	\centering
	\resizebox{.85\columnwidth}{!}
	{
		\smallskip
		\begin{tabular}{c|c|c|c|c}
			\hline
			~ & Method         & mAcc          & oAcc          & mIoU  \\
			\hline
			\multirow{6}{*}{5-th fold}
			  & PointNet       & 52.1          & 83.5          & 43.4  \\
			~ & SEGCloud       & 57.4          & -             & 48.9  \\
			~ & RSNet          & 59.4          & -             & 51.9  \\
			~ & 3P-RNN         & \textbf{71.3} & 85.7          & 53.4  \\
			~ & ASIS           & 60.9          & 86.9          & 53.4  \\
			~ & JSNet   (Ours) & 61.4          & \textbf{87.7} & \textbf{54.5} \\
			\hline\hline
			\multirow{6}{*}{6 fold}
		      & PointNet       & 60.3          & 80.3          & 48.9  \\
			~ & 3P-RNN         & \textbf{73.6} & 86.9          & 56.3  \\
			~ & MT-PNet	       & -             & 86.7          & -     \\
			~ & MV-CRF 	       & -             & 87.4          & -     \\
			~ & ASIS 	       & 70.1          & 86.2          & 59.3  \\
			~ & JSNet   (Ours) & 71.7          & \textbf{88.7} & \textbf{61.7} \\
			\hline
		\end{tabular}
	}
	\label{tab:s3dis_sem}
\end{table}

\subsection{ShapeNet Results}
\begin{figure}[t]
	\centering
	\includegraphics[width=0.9\columnwidth]{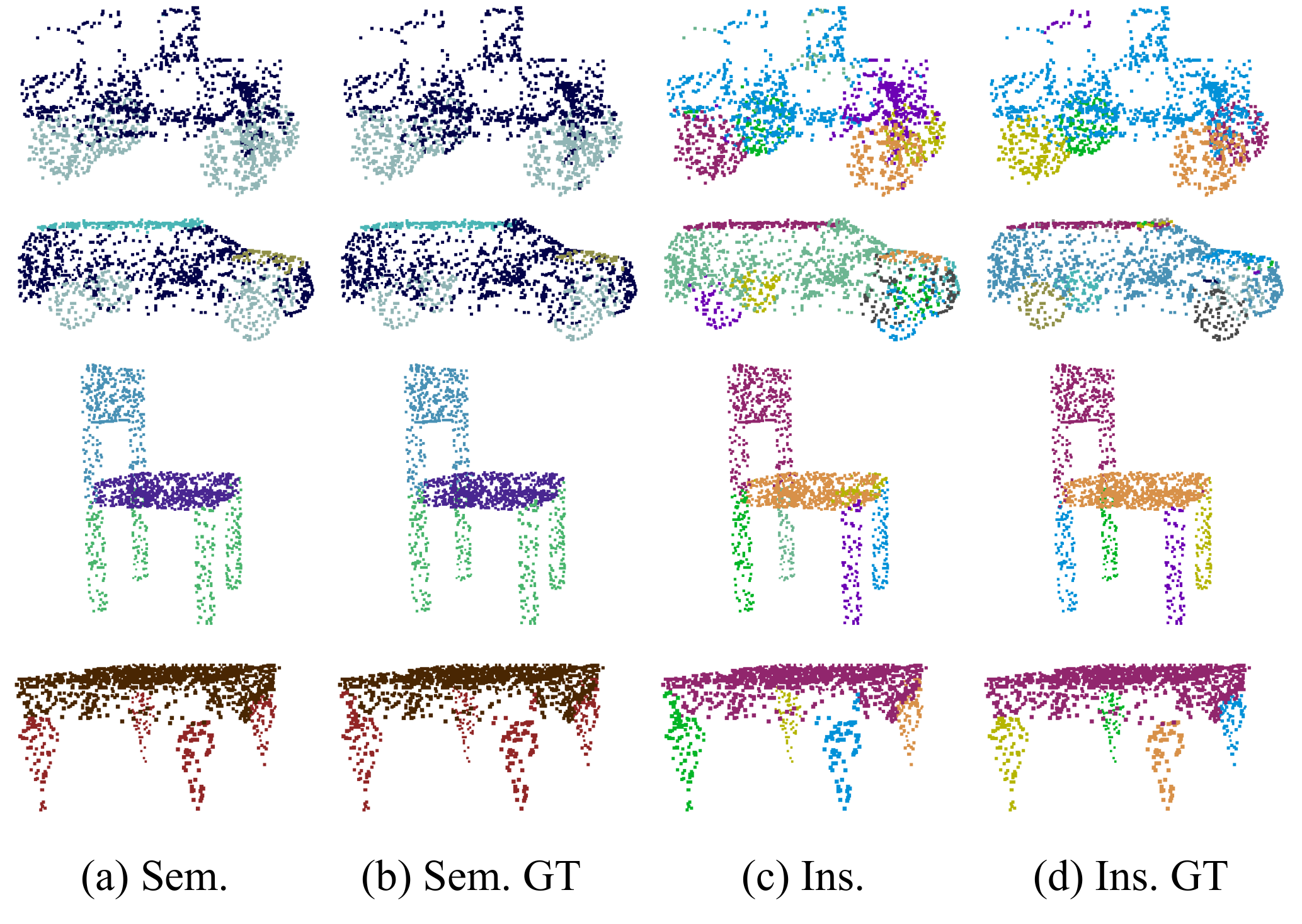}
	\caption{Qualitative results of JSNet on ShapeNet dataset. (a) Semantic prediction results of JSNet. (b) Ground truth of semantic segmentation. (c) Instance prediction results of JSNet. (d) Generated instance annotation for instance segmentation.}
	\label{fig:shapenet_results}
\end{figure}

Besides evaluation on the large scale indoor real scene benchmark S3DIS, we also conduct experiments on ShapeNet dataset. Following \cite{Wang_2018_CVPR}, the instance annotations are generated as ground truth to train our network. Since these annotations are fake ground truth labels, we only present the qualitative results of part instance segmentation, as is illustrated in Figure \ref{fig:shapenet_results}. The results of semantic segmentation are reported in Table \ref{tab:shapenet_sem}. We use PointNet++ \cite{qi2017pointnet++} as our baseline, and JSNet outperforms the baseline by 0.9-point mIoU. Compared with ASIS \cite{Wang_2019_CVPR}, our approach also achieve an improvement of 0.8 mIoU. These results show that our approach is also favorable for the part segmentation task.

\begin{table}[t]
	\caption{Semantic segmentation results on ShapeNet dataset.}
	\smallskip
	\centering
	\begin{tabular}{c|c}
		\hline
		Method         & mIoU  \\
		\hline
		PointNet       & 83.7  \\
		PointNet++     & 84.9  \\
		ASIS           & 85.0  \\
		JSNet (Ours)   & \textbf{85.8} \\
		\hline
	\end{tabular}
	\label{tab:shapenet_sem}
\end{table}

\subsection{Ablation Study}
\begin{table}[t]
	\caption{Ablation experiments results on Area 5 of the S3DIS. The short names for components and strategies are defined as: BN$-$Base network, BBN$-$Backbone network, IF$-$Instance fusion branch of JISS module, SF$-$Semantic fusion branch of JISS module, ES$-$Early stopping, RS$-$Random sample.}
	\smallskip
	\centering
	\resizebox{.95\linewidth}{!}
	{
		\smallskip
		\begin{tabular}{c|ccccc|cc|c|c}
			\hline
			\multirow{2}{*}{Group} & \multicolumn{5}{c|}{Component} & \multicolumn{2}{c|}{Strategy} & \multicolumn{2}{c}{Metric} \\
			\cline{2-10}
			~ & BN & BBN & PCFF & IF & SF & ES & RS & mPrec  & mIoU \\
			\hline
			(1) & $\surd$ & ~       & ~       & ~       & ~       & ~       & ~       & 52.3  & 52.7 \\
			(2) & ~       & $\surd$ & ~       & ~       & ~       & ~       & ~       & 55.9  & 53.0 \\
			(3) & ~       & $\surd$ & $\surd$ & ~       & ~       & ~       & ~       & 58.6  & 54.5 \\
			(4) & ~       & $\surd$ & ~       & $\surd$ & ~       & ~       & ~       & 56.9  & 53.5 \\
			(5) & ~       & $\surd$ & ~       & ~       & $\surd$ & ~       & ~       & 57.2  & 53.5 \\
			(6) & ~       & $\surd$ & ~       & $\surd$ & $\surd$ & ~       & ~       & 58.6  & 54.3 \\
			(7) & ~       & $\surd$ & $\surd$ & $\surd$ & $\surd$ & ~       & ~       & 57.6  & 54.3 \\
			(8) & ~       & $\surd$ & $\surd$ & $\surd$ & $\surd$ & $\surd$ & ~       & 58.7  & 54.4 \\
			(9) & ~       & $\surd$ & $\surd$ & $\surd$ & $\surd$ & ~       & $\surd$ & 62.1  & 54.5 \\
			(10) & ~      & $\surd$ & $\surd$ & $\surd$ & $\surd$ & $\surd$ & $\surd$ & \textbf{62.9} & \textbf{55.0} \\
			\hline
		\end{tabular}
	}
	\label{tab:s3dis_ablation_module}
\end{table}

To better validate the effectiveness of each component in our network, we conduct 7 groups of ablation experiments on Area 5 of S3DIS dataset. In addition, we also conduct additional experiments to validate the effects of different training strategies on the same dataset. For all ablation experiments, if there are no extra notes, we use the same configuration in the subsection Implementation Details. 

\noindent(1) \textbf{Base Network.} The base network includes a shared encoder and two parallel decoders. The encoder is built by stacking four set abstraction modules of PointNet++ \cite{qi2017pointnet++}, and the decoders are built by stacking four feature propagation modules of PointNet++.

\noindent(2) \textbf{Backbone Network.} The encoder of backbone is built by concatenating a set abstraction module of PointNet++ and three feature encoding layers of PointConv \cite{wu2018pointconv}. Similarly, the decoders are composed with three depthwise feature decoding layers of PointConv following a feature propagation module of PointNet++.

\noindent(3)-(6) \textbf{Single Module Evaluation.} We remove other components from the full framework (7) and only retain a module for the ablation experiments respectively.

\noindent(8)-(10) \textbf{Different strategies.} we train the full model with early stopping or random sample.

In Table \ref{tab:s3dis_ablation_module}, we present the ablation experimental results of different components in the full framework. 
Compared with the base network, the backbone network indeed benefits from a more efficient real convolution with density weighted. Compared with (2) and (3), the experimental results shows that fusing the feature of different layers could improve the segmentation precision because of the richer features after fusing. As for the only instance fusion semantic segmentation and only semantic awareness instance segmentation, the results indicate that better instance predictions could assign more reliable category labels to semantic branch, which can improve the performance of semantic segmentation. Similarly, the semantic awareness could enhance the instance predictions. In sixth ablation experiment, we combine instance fusion with semantic awareness, and the performance improvement is larger than only using one of them.

Table \ref{tab:s3dis_ablation_module} also depicts the ablation experiments results of the full framework trained with different schemes including fix sample, fix sample with early stopping, random sample and random sample with early stopping. Compared with (7) and (8), our model has a slight overfitting and we alleviate this phenomenon by training the network with early stopping. Compared the fix sample (7) with random sample (9), the results indicates that we train the full framework with the strategies random sample, which makes our model avoid overfitting and has stronger generalization ability. In addition, we also use the early stopping strategies to train the network with random sample. our approach achieves 62.9 mPrec and 55.0 mIoU for instance segmentation and semantic segmentation respectively.

\section{Conclusion}
In this work, we propose JSNet, which is a novel end-to-end approach based on deep learning framework for 3D instance segmentation and semantic segmentation on point clouds. The framework consists of a shared feature encoder, two parallel feature decoders followed by a point cloud feature fusion (PCFF) module respectively and a joint instance semantic segmentation (JISS) module. On the one hand, the feature encoder, the feature decoders and the PCFF module can learn more effective and more discriminative features. On the other hand, the JISS module make the instance and semantic segmentation take advantage of each other. Finally, our approach achieves a significant improvement in both instance and semantic segmentation tasks on S3DIS dataset. In the future, spatial geometric topology of point clouds can be added into our framework for better segmentation results.

\section{Acknowledgements}
This work was supported by the National Natural Science Foundation of China under Grants 61772213 and 91748204.

\bibliography{references}
\bibliographystyle{aaai}

\end{document}